\documentclass[sn-mathphys,Numbered]{sn-jnl}

\usepackage{multirow}%
\usepackage{amsmath,amssymb,amsfonts}%
\usepackage{amsthm}%
\usepackage{mathrsfs}%
\usepackage[title]{appendix}%
\usepackage{xcolor}%
\usepackage{textcomp}%
\usepackage{manyfoot}%
\usepackage{booktabs}%
\usepackage{algorithm}%
\usepackage{algorithmicx}%
\usepackage{algpseudocode}%
\usepackage{listings}
\usepackage{multicol,caption}%
\usepackage{hyperref}
\usepackage{commath}
\usepackage{graphicx}
\usepackage{comment}
\usepackage{adjustbox}

\theoremstyle{thmstyleone}%
\theoremstyle{thmstyletwo}%

\theoremstyle{thmstylethree}%

\begin{document}

\title[Adversarial Attacks and Dimensionality in Text Classifiers]{Adversarial Attacks and Dimensionality in Text Classifiers}


\author*[1]{\fnm{Nandish} \sur{Chattopadhyay}}\email{nandish001@e.ntu.edu.sg}

\author*[2]{\fnm{Atreya} \sur{Goswami}}\email{atreya@iitk.ac.in}

\author*[1]{\fnm{Anupam} \sur{Chattopadhyay}}\email{anupam@ntu.edu.sg}

\affil*[1]{\orgdiv{School of Computer Science and Engineering}, \orgname{Nanyang Technological University}, \orgaddress{ \country{Singapore}}}

\affil*[2]{\orgdiv{Computer Science and Engineering Department}, \orgname{Indian Institute of Technology Kanpur}, \orgaddress{ \country{India}}}


\abstract{Adversarial attacks on machine learning algorithms have been a key deterrent to the adoption of AI in many real-world use cases. They significantly undermine the ability of high-performance neural networks by forcing misclassifications. These attacks introduce minute and structured perturbations or alterations in the test samples, imperceptible to human annotators in general, but trained neural networks and other models are sensitive to it. Historically, adversarial attacks have been first identified and studied in the domain of image processing. In this paper, we study adversarial examples in the field of natural language processing, specifically text classification tasks. We investigate the reasons for adversarial vulnerability, particularly in relation to
the inherent dimensionality of the model. Our key finding is that there is a very strong correlation between the embedding dimensionality of the adversarial samples and their effectiveness on models tuned with input samples with same embedding dimension. We utilize this sensitivity to design an adversarial defense mechanism. We use ensemble models of varying inherent dimensionality to thwart the attacks. This is tested on multiple datasets for its efficacy in providing robustness. We also study the problem of measuring adversarial perturbation using different distance metrics. For all of the aforementioned studies, we have run tests on multiple models with varying dimensionality and used a word-vector level adversarial attack to substantiate the findings.}

\def\thefootnote{*}\footnotetext{This paper is accepted for publication at EURASIP Journal on Information Security in 2024.}



\maketitle

\section{Introduction}

Despite the boom in machine learning and AI usage in driving a wide range of applications, adversarial attacks remain a key deterrent to its adoption in practice. The vulnerability of high performance neural networks was demonstrated for the first time in 2015 ~\cite{szegedy}. The initial observations were made in the domain of computer vision in image processing tasks. Adversaries could generate slightly tweaked test samples, fooling the trained networks into misclassifications. These minute structured perturbations would be imperceptible to the human annotator and would have significant degradational effect on the performance of the model, with lots of misclassifications ~\cite{basic}. Other interesting properties, like the transferability of adversarial samples, were observed soon after ~\cite{papernot}. This led to the growth in the study of adversarial attacks by researchers. As a result, multiple adversarial attacks and corresponding defence mechanisms were proposed ~\cite{sir_survey}.


Naturally, as the threat of adversarial attacks undermined the trust stakeholders had in machine learning driven systems, researchers made multiple attempts to understand and explain this vulnerability. Some initial works in the domain attributed the origin of adversarial attacks to the linear nature of neural networks ~\cite{linear}. There were counterarguments made by other groups, particularly those studying the optimization landscape of neural networks and attributing the adversarial vulnerability to the properties of high dimensional spaces ~\cite{googledimension}. This area of work gained attention as the counter-intuitive properties of the geometry of high dimensional spaces could be empirically correlated to the observed properties of adversarial examples. Despite such efforts, adversarial attacks still pose a genuine threat to the reliable deployment of machine learning models in safety-critical applications, where errors could lead to catastrophes. A greater understanding of this vulnerability is therefore necessary.

\subsection{Motivation}

There are a lot of available studies on adversarial attacks on image classification models, primarily because that is where these attacks were first observed. The literature available concerning adversarial images is rich \cite{sir_survey}, with a continuous arms race between attacks and defences. The idea of introducing structured perturbations imperceptible to human annotators is agnostic of the use case and extends equally well in other domains, like natural language understanding. The methods are not directly possible to map, and natural language has its own distinctive properties and nuances in structure. This necessitates a thorough study of adversarial attacks in text modelling tasks. Neural architectures like recurrent neural networks are inherently different to convolutional neural networks used in image classification tasks, and therefore, so are the attacks and corresponding defence mechanisms. While there is some literature available on different attack schemes on text classifiers, our motivation is to try and understand the underlying reason that contributes to the success of the attacks. Specifically, we study how the dimensionality of the embedding vector space on top of which the models are built affects the adversarial vulnerability. There is a requirement to understand the impact of the size of the vector embedding model and, consequently, optimization landscape for the neural network on corresponding adversarial examples. This serves as the primary motivation of this paper, and we attempt to use the insights gained to propose a working defence mechanism against attacks. 

\begin{figure}[!htbp]
\centering
\includegraphics[width=\textwidth]{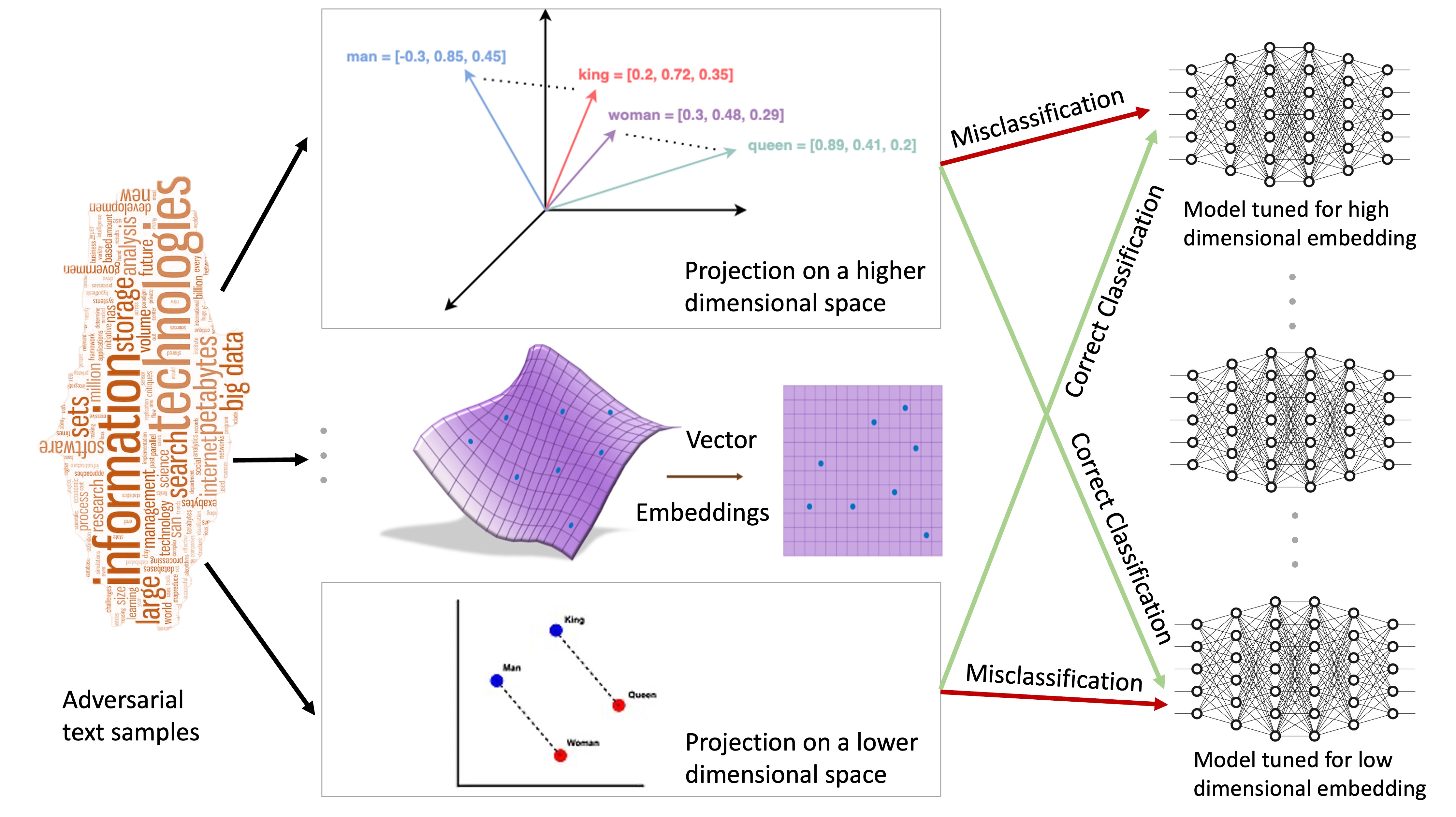}
\caption{Representation diagram of the sensitivity of the models' vulnerability on embedding dimensions for adversarial vulnerability.  }
\label{concept}
\end{figure}

We have identified a strong relationship between the adversarial vulnerability of the models and the corresponding dimension of embedding of the inputs using which the model was trained. The adversarial samples are very effective only when they are generated against a model which matches the samples' embedding dimension. This is intuitively represented in a concept diagram in Figure \ref{concept}. To the best of our knowledge, this is the first work in this domain, that relates the dimensionality of the word vector embedding to the adversarial vulnerability, in text classification tasks. Our proposed defense mechanism, based on an ensemble of models, is able to outperform the state-of-the-art in adversarial defences \cite{sota1} on similar tasks (using the same combination of model, dataset and attack). 

\subsection{Contribution}
The novel contributions of this paper are inclusive of, but not limited to, the following:
\begin{itemize}
    \item \textbf{Linking adversarial vulnerability to embedding dimension of inputs to the model:} Studying word-level adversarial attacks on text classifier neural architectures and understanding their behaviour with dimensionality. We have looked at the performance of the adversarial attacks on different models with different input embedding dimensions and optimization landscape thereof to establish the correlation, substantiated extensively through experimental results.
    \item \textbf{Using Ensemble models to thwart adversarial attacks:} Analyzing the sensitivity of the success of adversarial attacks on these models to develop potential defence mechanisms using ensembles of models. We have observed that adversarial attacks on text work well only when the target model's input embedding dimension matches that of the one for which the attack was created. We have used this fact to build robust ensembles with models having varying dimensionality to bypass the effect of adversarial attacks.
    \item \textbf{Measuring Adversarial Perturbation:} Studying the effect of different distance metrics on the measurement of adversarial perturbation. We have compared the performance of different distance metrics for the same. 
\end{itemize}

\subsection{Organization}
In this paper, Section \ref{background} outlines the relevant background and the related works in the domain of adversarial attacks on text classifiers. Then, Section \ref{adv_dimension} mentions the fundamental justification of intertwining dimensionality and adversarial examples and how they are generated. Section \ref{implementation} presents the details of the implementation used in this work. The experimental design, results, and key findings are presented in Section \ref{experiments}. We end with concluding remarks and the future scope of work in Section \ref{conclusions}. 

\section{Background and Related Works} \label{background}

In this section, we go through a brief overview of the literature available in the domain and outline the details of adversarial attacks in general and how they have been studied in the space of natural language processing tasks.

\subsection{Text Classifiers}
Text Classification can be defined as the machine learning technique used for the classification of a given text document under a predefined class. Suppose $d_i$ is a document in the entire arrangement $D$ of documents, and $\{c_1, c_2, ...., c_n\}$ is the set of classes, then after classification, class $c_i$ is assigned to the document $d_j$~\cite{rajan1}. Text classification is a crucial task in Natural Language Processing with many applications, for example, sentiment analysis, spam detection, topic labelling and intent detection. 

Machine Learning-based automatic text classification models learn different associations between tokens of the text and assign a particular output (i.e., class) to a particular input (i.e., text). The first step for training an NLP classifier is preprocessing of textual data. Preprocessing in the case of text dataset includes removing punctuation (.,!\$()*\%\@), removing URLs and lower casing the text~\cite{minaee2021deep}, tokenization~\cite{webster1992tokenization}, stopwords removal~\cite{kannan2014preprocessing} and stemming~\cite{kannan2014preprocessing}. 

\begin{figure}[!ht]
\centering
\includegraphics[width=\textwidth]{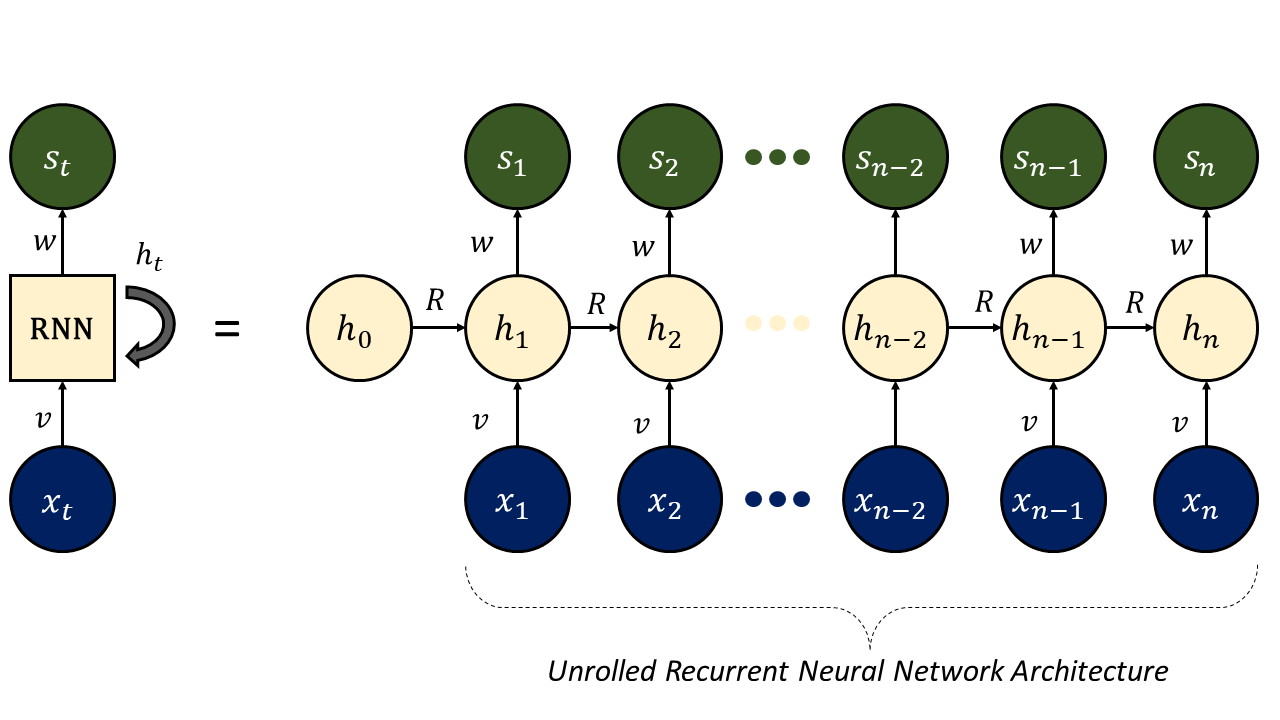}
\caption{A recurrent neural network architecture for text classification tasks. }
\label{rnn}
\end{figure}

Deep learning architectures provide a lot of advantages for text classification since they are inspired by how the human brain operates, called neural networks and can perform extremely well relative to other methods. Deep learning approaches such as Word2Vec~\cite{rong2014word2vec} or GloVe~\cite{pennington2014glove} have improved the effectiveness of classifiers learned with typical machine learning algorithms by obtaining better vector representations for words. Convolutional Neural Networks (CNN) and Recurrent Neural Networks (RNN) (illustrated in Figure \ref{rnn}) are the two basic deep learning architectures for text classification~\cite{yin2017comparative}. 

\subsection{Adversarial Examples}

Supervised machine learning, especially in specific tasks relating to image processing and natural language processing, has benefited strongly from the introduction of neural networks. Adversarial attacks were first observed in neural architectures built for image classification problems, too. For a particular classifier and a test data point that has been correctly classified by the network, a corresponding adversarial sample is a modified test sample which has been altered by the introduction of small structured perturbation that is not picked up by a human annotator. Adversarial attacks could be applicable to both computer vision and NLP models.

Typically, a text classification problem would consist of four key parts. First is the machine learning model, which is generally some neural architecture best suited for the task. Second, there needs to be a training dataset, which in practice is a sample picked from a population of all corpora combined. In the case of a supervised learning problem, like the one we are working on in this paper, the training dataset has associated labels. Third, a test dataset is there for testing the model's predictions. Fourth, once the model is trained over the dataset, the trained class manifolds are learnt within the space of the dataset, which splits it into the respective classes~\cite{bishop}.

\begin{figure}[!htbp]
\centering
\includegraphics[width=\textwidth]{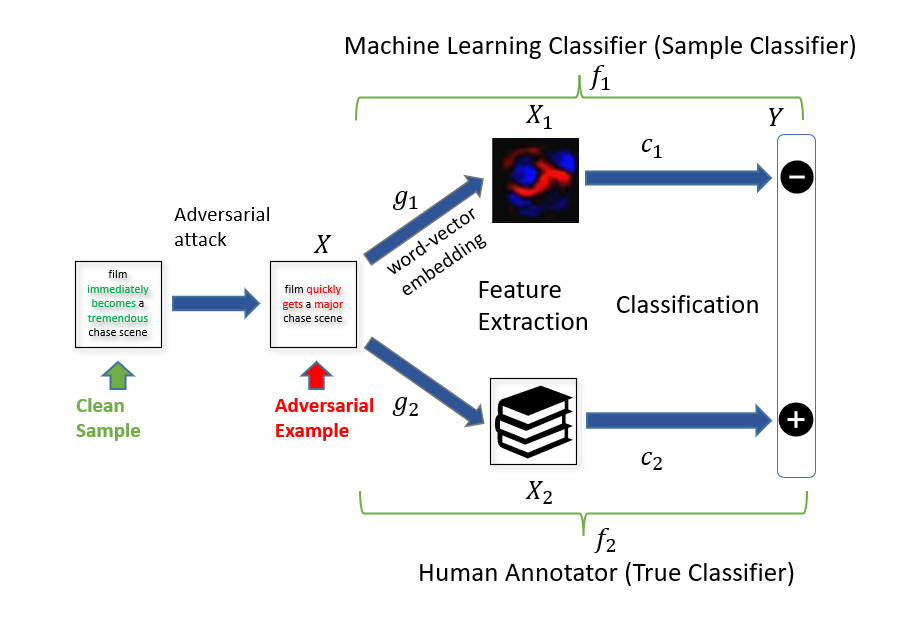}
\caption{An adversarial attack on a text classifier neural network model. }
\label{adv1}
\end{figure}

Considering a binary classification problem, a true classifier always correctly classifies the test samples, like a human annotator that sets the ground truth. If this had been known to us, a machine-learning model would not have been required, and a simple rule-based setup would have sufficed. As an approximation of the true classifier, we build a model that is able to learn the manifolds split by the classifier hyperplane over the landscape. Although the true classifier exists within the population and the learned model works in the sample, we reasonably assume that the true classifier also extends to the sample. Naturally, a notional gap exists between the true classifier and its approximation, trained on a non-exhaustive sample set. This leads to the creation of an adversarial space, as any sample that belongs to this space would lie on two different sides of the two classifiers mentioned earlier, and therefore lead to a disagreement on the prediction with respect to the machine learning model and the human annotator. Any sample lying in that space is, therefore, capable of demonstrating adversarial properties. Additionally, if many samples are present near the boundaries of the trained manifolds, then the introduction of a little structured perturbation can shift them across into the area that leads to the vulnerabilities. It is to be noted here that if the perturbation is beyond a specific threshold, the samples move across both classifiers, therefore not being adversarial any longer.

For the formal definition, following the diagram in Figure \ref{adv1}, let us assume that $X$ is the input sample space. To define the adversarial example, we consider two classifiers, as mentioned earlier, $f_1$ (sample classifier) and $f_2$ (human annotator). Each of these classifiers has separate components within them for extracting features and the actual task of classification. Then we have $X_1$ to be the feature space of the sample classifier, which in fact, is the word-vector embedded representation of the input sample. The dimensionality of this is the inherent dimension of the model, as per our definition. $X_2$ is the feature space used by the human annotator. Furthermore, we also have $d_1$ and $d_2$ to be distance metrics, norms which are respectively defined in $X_1$ and $X_2$.

If we consider $x \in X$ to be a clean sample, the corresponding adversarial sample created using an adversarial attack is $x^*$, with a norm $d_2$ in the space $X_2$ and a bound on adversarial perturbation $\delta > 0$, we have:

\begin{equation} 
\centering
\label{eq1}
\begin{split}
 f_1(x) \neq f_1(x^*) \quad \mbox{and} \quad f_2(x) = f_2(x^*)\\
 \mbox{such that} \quad d_2(g_2(x),g_2(x^*))< \delta \\  
\end{split}
\end{equation}

There are two important aspects to note from the above. By definition, the dimensionality of the feature space facilitates the generation of adversarial examples. Also, adversarial perturbation is bounded by a threshold in nature.

\subsection{Literature Review}
Early works on adversarial attacks mostly focused on image classification tasks \cite{szegedy}, \cite{kurakin2018adversarial}. But generating adversarial examples for text classification is a far more challenging task since text perturbation is far more easily recognizable by a human annotator than an image perturbation, owing to its discrete nature. However, growing security concerns led to active research in the field of adversarial attacks on NLP systems. Some of these attacks use character-level perturbations \cite{hotflip}, \cite{gao2018black}, while some use word-level perturbations \cite{li2016understanding}, \cite{liang2017deep}, \cite{feng2018right}. Though initial adversarial attacks were based on a brute-force approach to find perturbations, gradually, further research in this field led to the discovery of a more systematic and organized approach towards finding the \lq best\rq\:perturbation. The adversarial attack mechanisms proposed can be classified into two classes - white-box attacks, where model parameters, gradients and training data are known and are used to generate adversarial samples, and black-box attacks, where no information is available about the model to be attacked. In the field of white-box attacks, recently, some attack mechanisms have been developed which use rule-based synonym replacement and replacement by same parts-of-speech (POS) techniques. These have led to more natural adversarial samples \cite{alzantot2018generating}, \cite{ren2019generating}. Many efficient black-box attack techniques have also come up, like BAE \cite{garg2020bae}, which uses the BERT masked language model (MLM) to search for word replacements, and TEXTBUGGER \cite{li2018textbugger}, which sorts sentences in order of their importance and uses a scoring function to identify important words in the black-box setting. We take a step back from the cat-and-mouse chase of adversarial attacks and defences and study the possible reasons why these attacks exist, which is lacking in the available literature. 

Various adversarial defense techniques have also been designed over the years to address the vulnerability of real life AI systems in dealing with adversarial attacks \cite{defense_survey}. Broadly these techniques can be classified based on whether the defense mechanism is aware of the form of adversarial attack. Some defense techniques use the candidate list of words to be replaced to generate a set of adversarial data and incorporate it into model training dataset \cite{jin2020bert}, \cite{si2020better}, \cite{li2020bert}. In some techniques, a \emph{certified space} is constructed from the candidate list, such that any substitution falling in the certified space cannot perturb the model \cite{huang2019achieving}, \cite{jia2019certified}. If the candidate list is not accessible, gradient based adversarial training is often used to improve defense against adversarial attacks in the NLP domain \cite{cheng2019robust}, \cite{hotflip}. Another method, which is more popular in the image domain, is the strategy of adversarial purification using generative models \cite{samangouei2018defense}, \cite{lecun2006tutorial}, \cite{nie2022diffusion}. This technique has been explored in the NLP domain as well \cite{li2022text}, where input samples are purified by masking and masks prediction using pre-trained masked language models.

\section{Dimensionality and Adversarial Attack} \label{adv_dimension}
Neural networks are built to optimize parameters that operate in an extremely high dimensional space. It is interesting to study how dimensionality plays an important role in the behaviour of neural networks in dealing with test samples.

\subsection{Properties of Dimensionality}

\begin{figure}[!htbp]
\centering
\includegraphics[width=\textwidth]{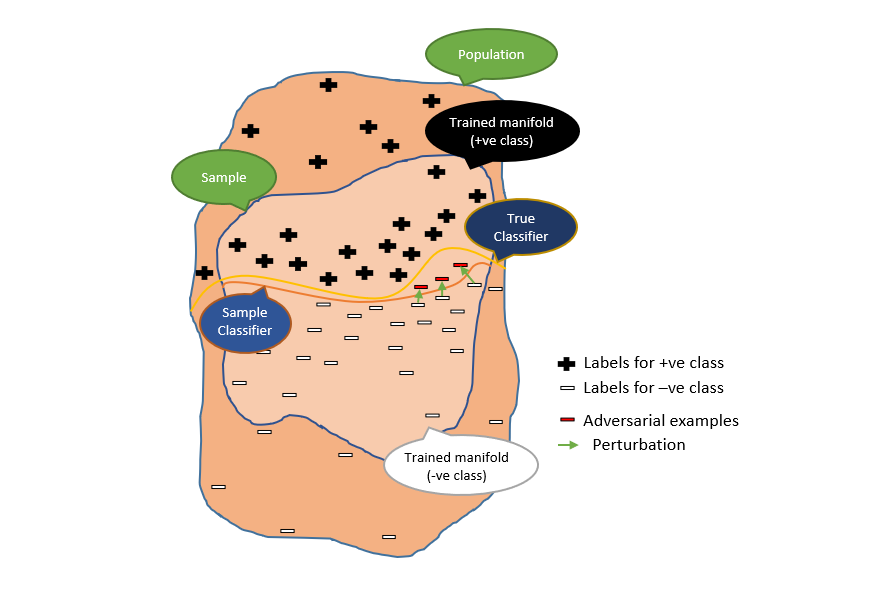}
\caption{Dimensionality and adversarial attacks.  }
\label{adv}
\end{figure}

The behaviour of data points residing in a high dimensional space is counter-intuitive because, irrespective of the distribution from which they are sampled, the spread is not even. There is a theoretical justification for most data points to exist close to the boundaries of the trained manifolds, away from the centres near the surface \cite{hopcroft}. Although this is not enough to rigorously establish causality, there is a notion that the generation of adversarial examples is facilitated by the existence of many sample points near the decision boundaries, as a small perturbation can transfer them across. This phenomenon is presented visually in Figure \ref{adv}.

There is a significant difference in the behaviour of data points in a lower dimensional space to that in a higher dimensional space. This is established theoretically, and it forms a key part of the explanation behind the sensitivity of the adversarial examples in different embedding spaces of different dimensions. Here, we showcase the theoretical formulation for the behaviour of the data points in the two most common data distributions.

\textsc{Uniform Distribution:} 
Contrary to the intuitive idea of any distribution in lower dimensions, for high dimensional settings, the majority of the $d$-dimensional unit object is present in a relatively very small annulus of width $O(1/d)$ near its periphery. If one generalises it to a $d$-dimensional ball with radius $r$, the corresponding annulus would be of width of $O(r/d)$. Interestingly, the majority of the volume of the data points in the distribution not only lies in an annulus, but is also highly concentrated near the equator, in a ring.Specifically, one can demonstrate that at least a $1-\frac{2}{c}e^{-c^2/2}$ fraction of the volume of the $d$-dimensional unit ball has $\abs{x_1} \leq \frac{c}{\sqrt{d-1}}$, for any $c \geq 1$ and $d \geq 3$.

\textsc{Gaussian Distribution:}
Generally, the regular $1$-dimensional Gaussian distribution is such that most of the volume of data points are concentrated near the mean. 
According to the \emph{Gaussian Annulus Theorem}~\cite{hopcroft}, the spherical Gaussian distribution of variance unity in $d$-dimensions has all but at most $3e^{-c \beta^2}$ of the probability mass lying within the annulus $\sqrt{d} - \beta \leq \abs{x} \leq \sqrt{d} + \beta$, for any $\beta \leq \sqrt{d}$ and $c$ being a fixed positive constant.

Often, the properties of the high dimensional space are attributed to be one of the primary contributors to adversarial attacks being successful \cite{chattopadhyay}. This has been well-tested in the domain of computer vision.  One of the major objectives of this work is to investigate empirically how the behaviour of adversarial attacks on text classifiers depends on the inherent dimensionality of the text classification problem.

\subsection{Dimension Sensitivity} \label{pp}
The dimensionality of adversarial attacks is essentially related to the vector embedding dimension of the input to the model against which the attack is mounted. The input to the text classifier networks in most applications of natural language processing is a bunch of vector representations that have been mapped to a learned vector space over the corpus. The size of the input vectors naturally depends on the number of elements considered to create that vector embedding space, which is a user choice when the model is trained. The optimization problem rests on this set of vectors. During training, the parameters of the neural network, which are the weight matrices, are tuned via some optimization technique like gradient descent etc., and the individual gradients are stored. The trained manifolds that split the landscape into the corresponding classes are separated by the hyperplane classifier. 

When the attacks are introduced to clean samples, they are always implemented with respect to some trained model, which it aims to fool. This relates the embedding dimensionality of the model input to the attack mechanism. The attack tries to search and make perturbations to the samples to shift the adversarial example across the separating hyperplane, residing within the optimization landscape characterized by the embedding input dimension. It is, therefore, natural to expect that these adversarial examples will be sensitive to dimensionality. We have studied to what extent the embedding dimensionality affects adversarial vulnerability and how a change in the dimensionality can bring about failures in the attack, which, in fact, could lead us to a potential defence. 

As demonstrated by thorough experimentation in Section \ref{experiments}, there is a strong dependency between the success of the attack and the inherent dimensionality. This is quite unlike the observations made in the domain of images, where higher dimensionality facilitates adversarial attacks \cite{chattopadhyay}, but it is not true that the attacks are successful only for the models whose dimensionality matches with the attack's ones.

\subsection{Ensembling }
A key insight that comes out of the study of adversarial attacks and their inherent dimensionality is that the dependence can be used to build defence mechanisms. Adversarial examples in text samples are highly tuned with respect to the model, which it tries to fool and does not transfer well to other models trained on different embedding dimensions. This gives rise to a potential way of blocking adversarial examples by using an ensemble of models trained on different embedding dimensions. A majority vote could be used to determine the corresponding classes of the specific sample in question. The intuition is that the model whose dimensionality matches that of the adversarial sample would most likely misclassify the sample, but the other models will not. It is expected that the rest of the models will not suffer from adversarial performance degradation. In order for this idea to work, it is necessary to study the extent of dependency between adversarial examples and dimensionality, pertaining to how small a change in embedding dimension can lead to its failure. We have thoroughly tested the aforementioned scheme with multiple ensemble model setups to check for adversarial defence mechanisms.

\subsection{Measuring Adversarial Perturbation}
One key aspect of the generation of adversarial examples is that the adversarial perturbation has to be bounded, that is, a structured alteration of the samples, small enough to be imperceptible to the human annotator but significant enough to force a misclassification by a trained neural network. In order to establish the extent of modifications that can be made to the clean samples, it is necessary to measure adversarial perturbation.

The task of measuring adversarial perturbation is not easy, especially in a high-dimensional setting. The formal definition of adversarial examples in Eq.~\eqref{eq1} states that the adversarial perturbation is bounded by a predefined threshold $\delta$, which is measurable using some norm distance metric that is defined in the corresponding space. Since this work is studying the relevance of dimensionality in adversarial attacks, it is necessary to investigate the amount of adversarial perturbation required to create adversarial samples at different embedding dimensions. In the case of images, adversarial attacks are facilitated by higher dimensions, and most distance metrics do not provide meaningful measures of distances due to statistical properties ~\cite{metric}. We have studied how this notion holds in the context of text classifiers.

\section{Implementation} \label{implementation}
In this section, we discuss the details of the implementation of the adversarial attacks on the text classifier models, mentioning the design choices and the pipeline used. 

\subsection{Pipeline}
First, we present our experimental setup used to achieve the mentioned results. The datasets mainly used in all our experiments are the IMDB Movie Reviews dataset \cite{maas-EtAl:2011:ACL-HLT2011} and the Twitter Sentiment140 dataset \cite{go2009twitter}. The input pipeline takes as input the datasets in CSV file format. The datasets consist of 2 columns - review and sentiment, which may take values ``positive'' or ``negative''. Then the input data is preprocessed by encoding the sentiment as \lq 1\rq\:(positive) or \lq 0\rq\:(negative), encoding emojis, removing common stopwords from the review and removing data points with missing sentiment labels. The preprocessed data is fed to the embedding pipeline, where the words in the reviews are tokenized and converted into their corresponding word vector representations of a given vector size using the open-source Python library Gensim \cite{rehurek2011gensim}.

\textbf{Text Classifier Model:} An embedding matrix is created from the word embeddings and passed on to the training pipeline as a non-trainable embedding layer. The input data is divided into train, validation and test sets and fed to the LSTM classifier model. All the models are trained and tested using the open-source ML library Tensorflow \cite{tensorflow2015-whitepaper}. The model is compiled using Adam optimizer, binary cross-entropy loss and uses accuracy as a metric. The trained model and test dataset are then fed to the testing pipeline, which outputs the number of successful predictions by our trained model. 

\textbf{Adversarial Attack:} Moving on to the adversarial pipeline, it takes as input the trained model, the pure dataset, word embeddings and the tokenizer. The pipeline outputs the adversarial samples corresponding to the pure samples if the sample is correctly classified by our classifier.

\textbf{Robust Ensemble Models:} We have used two different types of ensemble models, comprising three and five models - one of which uses majority voting, and the other is a weighted averaged model, where the votes for the class decision are weighted by the respective output probabilities. It is to be noted here that each of the members of the ensemble has been trained with a different input embedding dimension.

\subsection{Models}

We considered two classifier models for the experiments. Our first classifier model is a sequential model which consists of an embedding layer, bidirectional LSTM layers and one or more Dense layers. Specifically, the sequential model for IMDB Movie Reviews sentiment classification consists of an embedding layer, an LSTM layer followed by a Dense layer, whereas the Twitter Sentiment Classification model consists of an embedding layer, two bidirectional LSTM layers and successive Dense layers with output vector dimensions 128, 64, 16 and 1. The Word2Vec algorithm is used to generate embedding vectors by training on the entire corpus. The embedding matrix thus generated is then used to initialize the embedding layer, which takes as input the sequence of words padded to the maximum length of the reviews. The LSTM layers take as input the sequence of word vectors generated by the embedding layer and output a 64-dimensional vector. This layer is accompanied by a dropout ratio of 0.2 and an L2 kernel regularizer. The output layer is a Dense layer that uses the sigmoid activation function and is supplemented with the L2 kernel regularizer. It outputs the predicted class label. Both the models are compiled using the Adam optimizer and binary cross-entropy loss function and trained for 25 epochs with a batch size of 64, with callbacks for reducing the learning rate on reaching a plateau in the optimization landscape and early stopping. We have also used a transformer model, which is widely used in natural language understanding tasks. Specifically, the one used in the experiments for benchmarking is the distilled version of a BERT base model - DistilBERT base model (uncased) \cite{sanh2019distilbert}. Essentially, it is a pre-trained model on large datasets of the English language through a self-supervised process that can be fine-tuned for specific language processing tasks. In general, the BERT-based model has two different objectives that it is able to fulfil, Masked Language Modelling and Next Sequence Prediction. The model is able to learn meaningful inner representations of the language, which is useful for many later tasks like text classification etc. For our experiments, we have fine-tuned the model using both datasets for 5 epochs with a batch size of 16 and a learning rate of $3\times 10^{-5}$.

\subsection{Adversarial Attack}
To study the impact of dimensionality on adversarial attacks, we used a word-level attack scheme that directly relates to the embedding dimension of the text classifier model. This enabled us to have more control over the process of varying and tuning the dimensionality settings for experiments. Our attack is roughly based on the TextFooler attack \cite{jin2020bert}, an adversarial attack mechanism that generates syntactically and semantically similar adversarial samples. The adversarial attack is designed in a black box setting i.e. the attack mechanism does not have access to the attacked model's architecture or parameters. It can only use queries made on the model, the resulting predictions and confidence scores to generate adversarial samples. 

The adversarial attack algorithm assigns \emph{Word Importance Ranking} to each word in the input. Difference between the original prediction of the model on the input and the prediction on the input after deleting a given word is used to calculate the ranking of the word. Thus, in order to alter the model prediction with minimal alterations, a greedy selection mechanism is used to select the word with maximum importance to be modified first. Once we identify the word with maximum importance, we gather a set of closest synonyms of the word in the defined vocabulary using cosine similarity to ensure maximum semantic similarity with the original word. To represent the words, we use counter-fitting word vectors \cite{mrkvsic2016counter} which are specially designed to represent synonymous words. Also, to ensure sentence semantic similarity between the original input sample and the generated adversarial sample, we use the Universal Sentence Encoder \cite{cer2018universal} to derive cosine similarity. After the candidate list is finalised for the word with the highest importance, we replace the word with one of the candidates if it alters the prediction of the model. If not, we replace the word with the candidate word which minimizes the confidence score of the original label. Then we repeat the procedure for the next most important word, until the model prediction is altered. The adversarial attack algorithm was implemented using the open-source Python library Textattack \cite{textattack}.

\section{Experiments} \label{experiments}
In this section, we present the details of the thorough experiments that have been run as part of our analysis. We have tested the aspect of dimensionality and its impact on adversarial attacks in text classifiers, along with studying the scope of potential defence mechanisms and measurement of adversarial perturbation. The experiments have been carried out on Microsoft Azure Machine Learning Studio \cite{barnes2015azure}. We used a Standard\_NC6 virtual machine (NVIDIA Tesla K80 GPU, 6 cores, 56GB RAM) to parallelize and accelerate the attack pipeline.

\subsection{Design}
The setup of the experiments is described here. We have studied:
\begin{itemize}
    \item \textbf{Relating adversarial vulnerability and dimensionality:} The adversarial vulnerability of text classifier neural networks has been studied for varying associated dimensionality, which is the embedding dimension of the input to the trained neural network upon which the attack has been established. Specifically, we observed the sensitivity of the adversarial attacks to the inherent dimensionality by using a telescopic approach to test how closely they are related. Small incremental changes were introduced to the embedding dimension of the models on which the attacks were mounted, and then the generated adversarial examples were tested on the rest of the models, as shown in Table \ref{table2}.
    \item \textbf{Adversarial robustness using ensembles:} We have used ensemble models to thwart the attacks, taking advantage of the fact that they work well only when the adversarial examples are subjected to a model whose input embedding dimension matches that of the one upon which the adversarial attack was created in the first place, as shown in Tables \ref{table1} and \ref{table2}. In this case, we have used multiple choices of ensemble models, with three and five parallel models and carried out a majority voting or weighted averaging afterwards on the output classes to assign a label to any specific sample in question. 
    \item \textbf{Measuring adversarial perturbation:} Considering the vectorized version of the samples (using word vector embedding of specific dimensions), for both clean and adversarial, we used different metrics to study the point-to-point distances. In order to better understand if the choice of metrics makes any difference or not in calculating the adversarial perturbations, we investigated the distribution of the distances for $100$ pairs of clean and adversarial samples. We looked at 5-point statistics to compare the distributions in Table \ref{table3}. The distance metrics used are the $L_1$, $L_2$ and $L_{\infty}$ norms. 
\end{itemize}

\subsection{Results}
The experimental results are presented in a tabular form in this section. Throughout this section, we use ``Dim: \{DIM\}'' to denote the Sequential model with embedding vector dimension \lq DIM\rq, ``\{x\}-Ens'' to denote a majority voting-based ensemble with \lq x\rq\:parallel models and ``\{x\}-Wt Ens'' to denote a weighted averaging based ensemble with \lq x\rq\:parallel models. The results corresponding to these two classes of models have been separated into two groups within each of the tables. 

Table \ref{table0} presents the basic benchmarking of models used in the experimental analysis against different embedding dimensions, their ensembles and transformers on the IMDB and Twitter datasets for baseline accuracies on the text classification task. All numbers presented in the table are classification percentage accuracies of the models on the test dataset.

\begin{table}[!htbp]
\centering
\caption{Benchmarking models used in the experimental analysis against different embedding dimensions and ensembles and transformers on the IMDB and Twitter datasets. }
\label{table0}
\begin{tabular}{l|cc}
\hline
\textbf{Model Type / Dataset}  & IMDB Dataset & Twitter dataset \\ \hline
Dim: 100       & 94.00        & 78.60           \\
Dim: 200       & 93.00        & 79.40           \\
Dim: 300       & 91.00        & 80.00           \\
Dim: 400       & 93.00        & 81.20           \\
Dim: 500       & 91.00        & 80.80           \\
Dim: 900       & 94.00        & 80.80           \\
Dim: 950       & 92.00        & 80.80           \\
Dim: 1000      & 95.00        & 81.80           \\
Dim: 1050      & 93.00        & 82.00           \\
Dim: 1100      & 93.00        & 79.80           \\ \hline 
3-Ens               & 95.00        & 80.80           \\
5-Ens               & 97.00        & 80.80           \\
3-Wt Ens      & 95.00        & 80.80           \\
5-Wt Ens      & 97.00        & 80.80           \\
DistilBERT (Transformer model) & 95.00        & 82.50             \\ \hline
\end{tabular}
\end{table}

Table \ref{table1} shows the study of the adversarial vulnerability of neural networks with varying inherent dimensionality, along with the performance of the ensemble models as a counter-measure. Each row corresponds to the model with the specified embedding vector dimension or an ensemble, and each column corresponds to the set of adversarial samples generated by an adversarial attack on the model with the specified embedding vector dimension. We observe the adversarial vulnerability of the models and the robustness of the ensembles. This set of results corresponds to the IMDB dataset. All numbers presented in the table are classification percentage accuracies of the models on the test dataset. Note that the strength of the adversarial attack is significantly higher when the model's embedding dimension matches the embedding dimension of the adversarial attack. The results for the most successful attacks are presented in bold.

\begin{table}[!htbp]
\centering
\caption{Study of the sensitivity of adversarial examples to inherent dimensionality and the use of ensemble models as a counter-measure, tested on the IMDB dataset. We present the evidence of the vulnerability of individual models in the first part and the effectiveness of the ensemble models in the later. }
\label{table1}
\begin{tabular}{l|cccccccccc}
\hline
\textbf{Model/Attack} & 100            & 200            & 300            & 400            & 500            & 900            & 950            & 1000           & 1050           & 1100           \\ \hline
Dim: 100                    & \textbf{26.00} & 76.00          & 78.00          & 81.00          & 79.00          & 81.00          & 79.00          & 78.00          & 79.00          & 80.00          \\
Dim: 200                    & 71.00          & \textbf{33.00} & 76.00          & 78.00          & 79.00          & 81.00          & 81.00          & 76.00          & 84.00          & 75.00          \\
Dim: 300                    & 71.00          & 68.00          & \textbf{29.00} & 79.00          & 77.00          & 76.00          & 78.00          & 74.00          & 82.00          & 75.00          \\
Dim: 400                    & 73.00          & 77.00          & 74.00          & \textbf{30.00} & 80.00          & 81.00          & 85.00          & 75.00          & 86.00          & 80.00          \\
Dim: 500                    & 69.00          & 70.00          & 75.00          & 76.00          & \textbf{27.00} & 76.00          & 75.00          & 72.00          & 80.00          & 74.00          \\
Dim: 900                    & 71.00          & 73.00          & 75.00          & 76.00          & 84.00          & \textbf{24.00} & 82.00          & 74.00          & 83.00          & 77.00          \\
Dim: 950                    & 73.00          & 72.00          & 77.00          & 76.00          & 76.00          & 75.00          & \textbf{28.00} & 69.00          & 81.00          & 74.00          \\
Dim: 1000                   & 74.00          & 76.00          & 82.00          & 81.00          & 79.00          & 82.00          & 80.00          & \textbf{22.00} & 79.00          & 85.00          \\
Dim: 1050                   & 71.00          & 78.00          & 77.00          & 73.00          & 79.00          & 77.00          & 72.00          & 75.00          & \textbf{23.00} & 74.00          \\
Dim: 1100                   & 82.00          & 75.00          & 79.00          & 78.00          & 76.00          & 82.00          & 78.00          & 76.00          & 81.00          & \textbf{28.00} \\ \hline 
3-Ens                       & 65.00          & 78.00          & 84.00          & 82.00          & 68.00          & 86.00          & 83.00          & 64.00          & 83.00          & 85.00          \\
5-Ens                       & 68.00          & 73.00          & 76.00          & 82.00          & 82.00          & 75.00          & 82.00          & 75.00          & 83.00          & 75.00          \\
3-Wt Ens              & 65.00          & 78.00          & 84.00          & 82.00          & 68.00          & 86.00          & 83.00          & 64.00          & 83.00          & 85.00          \\
5-Wt Ens              & 68.00          & 73.00          & 77.00          & 82.00          & 83.00          & 78.00          & 82.00          & 75.00          & 83.00          & 75.00          \\ \hline
\end{tabular}%
\end{table}

Table \ref{table2} presents the study similar to the earlier results, with the exception of this set of results being tested on the Twitter dataset. All numbers presented in the table are classification percentage accuracies of the models on the test dataset. Note that the strength of the adversarial attack is significantly higher when the model's embedding dimension matches the embedding dimension of the adversarial attack. The results for the most successful attacks are presented in bold.

\begin{table}[!htbp]
\centering
\caption{Study of the sensitivity of adversarial examples to inherent dimensionality and the use of ensemble models as a counter-measure tested on the Twitter dataset. We present the evidence of the vulnerability of individual models in the first part and the effectiveness of the ensemble models in the later. }
\label{table2}
\begin{tabular}{l|cccccccccc}
\hline
\textbf{Model/Attack} & 100            & 200            & 300            & 400            & 500            & 900            & 950            & 1000           & 1050           & 1100           \\ \hline
Dim: 100          & \textbf{56.40} & 65.00          & 68.20          & 68.40          & 70.60          & 69.80          & 71.40          & 70.40          & 72.60          & 71.80          \\
Dim: 200          & 70.40          & \textbf{58.60} & 68.60          & 70.80          & 73.60          & 71.00          & 73.20          & 71.20          & 73.20          & 73.80          \\
Dim: 300          & 71.20          & 69.00          & \textbf{59.40} & 70.60          & 72.00          & 72.40          & 74.40          & 72.40          & 73.40          & 74.00          \\
Dim: 400          & 72.60          & 69.80          & 68.20          & \textbf{63.00} & 71.20          & 72.00          & 73.60          & 73.00          & 74.40          & 73.80          \\
Dim: 500          & 73.00          & 69.60          & 69.00          & 71.40          & \textbf{63.00} & 71.60          & 71.80          & 72.20          & 73.40          & 71.80          \\
Dim: 900          & 71.20          & 69.80          & 71.20          & 71.60          & 72.00          & \textbf{63.80} & 73.40          & 71.40          & 71.60          & 70.80          \\
Dim: 950          & 74.00          & 70.60          & 71.20          & 72.40          & 71.20          & 72.00          & \textbf{64.60} & 71.80          & 72.40          & 73.20          \\
Dim: 1000         & 74.80          & 72.00          & 71.80          & 72.20          & 71.60          & 72.20          & 72.40          & \textbf{66.00} & 73.60          & 73.60          \\
Dim: 1050         & 73.80          & 72.20          & 71.60          & 71.80          & 71.80          & 71.20          & 72.80          & 72.00          & \textbf{67.00} & 73.60          \\
Dim: 1100         & 73.20          & 70.40          & 69.60          & 71.40          & 70.20          & 71.40          & 71.80          & 71.40          & 71.60          & \textbf{65.40} \\ \hline 
3-Ens              & 71.40          & 70.00          & 70.80          & 70.20          & 69.00          & 72.00          & 72.60          & 70.00          & 72.20          & 73.40          \\
5-Ens              & 70.60          & 69.00          & 67.80          & 71.40          & 71.00          & 71.00          & 73.80          & 72.00          & 73.60          & 72.00          \\
3-Wt Ens           & 71.40          & 70.00          & 70.80          & 70.20          & 69.00          & 72.00          & 72.60          & 70.00          & 74.40          & 73.40          \\
5-Wt Ens           & 70.60          & 69.00          & 68.00          & 71.80          & 71.00          & 71.00          & 74.00          & 72.00          & 73.60          & 72.00          \\ \hline
\end{tabular}
\end{table}

Table \ref{table3} presents the study of the measurement of adversarial perturbation with different metrics. The distance metrics used are the $L_1$, $L_2$ and $L_{\infty}$. The corresponding statistics of the distribution of distances between clean and adversarial samples are presented for different embedding dimensions. This set of results corresponds to the IMDB dataset.

\begin{table}[!htbp]
\centering
\caption{Measurements of Adversarial Perturbation using distance metrics against different embedding dimensions for the IMDB dataset.  }
\label{table3}
\begin{tabular}{c|c|ccccc}
\hline
Embedding      & Norm     & Min     & Median  & Max     & Std. Dev & CV      \\ \hline \hline  
                 & $L_{1}$    & 65.955  & 178.199 & 276.691 & 30.690   & 17.39\% \\
Dimension: 200  & $L_{2}$    & 5.826   & 15.878  & 24.121  & 2.718    & 17.39\% \\
                 & $L_{\infty}$ & 1.230   & 3.393   & 5.205   & 0.588    & 17.80\% \\ \hline
                & $L_{1}$    & 114.141 & 246.619 & 325.624 & 41.157   & 16.96\% \\
Dimension: 400  & $L_{2}$    & 7.324   & 15.549  & 20.940  & 20.940   & 16.89\% \\
                 & $L_{\infty}$ & 1.297   & 2.672   & 4.963   & 4.963    & 19.89\% \\ \hline
                & $L_{1}$    & 134.588 & 346.333 & 346.333 & 70.404   & 20.52\% \\
Dimension: 900  & $L_{2}$    & 5.660   & 14.774  & 22.185  & 3.017    & 20.65\% \\
                 & $L_{\infty}$ & 0.870   & 2.018   & 4.344   & 0.488    & 24.07\% \\ \hline
                & $L_{1}$    & 64.041  & 387.795 & 534.375 & 68.763   & 18.24\% \\
Dimension: 1100 & $L_{2}$    & 2.508   & 15.148  & 21.061  & 2.698    & 18.19\% \\
                 & $L_{\infty}$ & 0.323   & 2.111   & 3.337   & 0.451    & 21.81\% \\ \hline
\end{tabular}
\end{table}

Table \ref{table4} presents the results similar to that of the earlier table, the only difference being that these correspond to the Twitter dataset.

\begin{table}[!htbp]
\centering
\caption{Measurements of Adversarial Perturbation using distance metrics against different embedding dimensions for the Twitter dataset.  }
\label{table4}
\begin{tabular}{c|c|ccccc}
\hline
Embedding       & Norm     & Min    & Median  & Max     & Std. Dev & CV      \\ \hline \hline
                & $L_{1}$    & 13.820 & 30.951  & 51.307  & 5.681    & 18.38\% \\
Dimension: 200  & $L_{2}$    & 1.232  & 2.737   & 4.615   & 0.502    & 18.38\% \\
                & $L_{\infty}$ & 0.235  & 0.563   & 1.106   & 0.118    & 20.60\% \\ \hline
                & $L_{1}$    & 21.168 & 56.121  & 88.320  & 11.826   & 21.25\% \\
Dimension: 400  & $L_{2}$    & 1.307  & 3.515   & 5.542   & 0.743    & 21.29\% \\
                & $L_{\infty}$ & 0.159  & 0.553   & 1.023   & 0.132    & 23.79\% \\ \hline
                & $L_{1}$    & 24.602 & 100.080 & 153.294 & 25.300   & 25.76\% \\
Dimension: 900  & $L_{2}$    & 1.039  & 4.185   & 6.439   & 1.057    & 25.72\% \\
                & $L_{\infty}$ & 0.120  & 0.480   & 0.794   & 0.124    & 26.41\% \\ \hline
                & $L_{1}$    & 27.289 & 109.371 & 171.712 & 28.813   & 26.94\% \\
Dimension: 1100 & $L_{2}$    & 1.046  & 4.137   & 6.480   & 1.087    & 26.85\% \\
                & $L_{\infty}$ & 0.109  & 0.436   & 0.705   & 0.122    & 28.13\% \\ \hline
\end{tabular}
\end{table}
Note that the space complexity of the ensemble models is $k$ times that of the individual sequential model trained on a particular embedding vector dimension, where $k$ is the number of models constituting the ensemble. Time complexity of the ensemble will be of the same order as that of an individual model, as the models are run in parallel.

\subsection{Key Findings}
The most important findings and key takeaways are mentioned here:
\begin{itemize}
    \item We observe that the adversarial attacks are highly successful when they are subjected to models which have the same input embedding dimension as the one with respect to which the set of adversarial samples was generated. This is seen in Figures  \ref{performance_1} and \ref{performance_2}, where the bars are low only when they correspond to the scenario where the model's embedding dimension matches the dimension for which the attack was created. The rest of the bars are relatively much higher. The bars, representing performance accuracies, support the assertion about adversarial sensitivity towards inherent dimensionality. Since the adversarial attacks are generated using a fixed embedding dimension, which is the vector space from which the replacement words are chosen, we observe that changing the embedding dimension of the model makes the same adversarial attack weak. 
    \begin{figure}[!htbp]
    \centering
    \includegraphics[width=\textwidth]{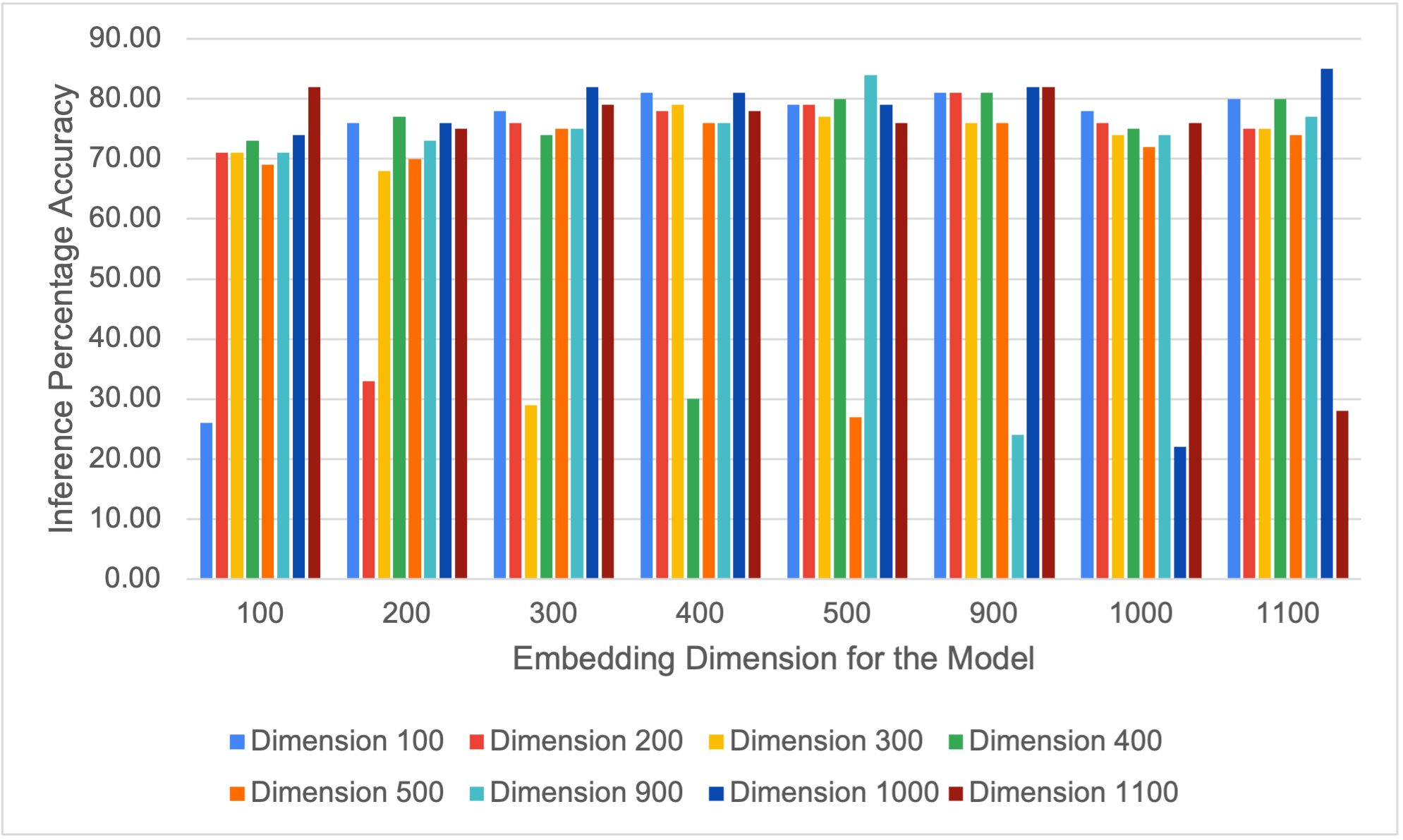}
    \caption{Adversarial vulnerability is sensitive to the embedding dimension, tested on the IMDB dataset. Note that for every model, only one attack works (lower bar), when its embedding dimension matches that of the model's inputs. }
    \label{performance_1}
    \end{figure}
    \begin{figure}[!htbp]
    \centering
    \includegraphics[width=\textwidth]{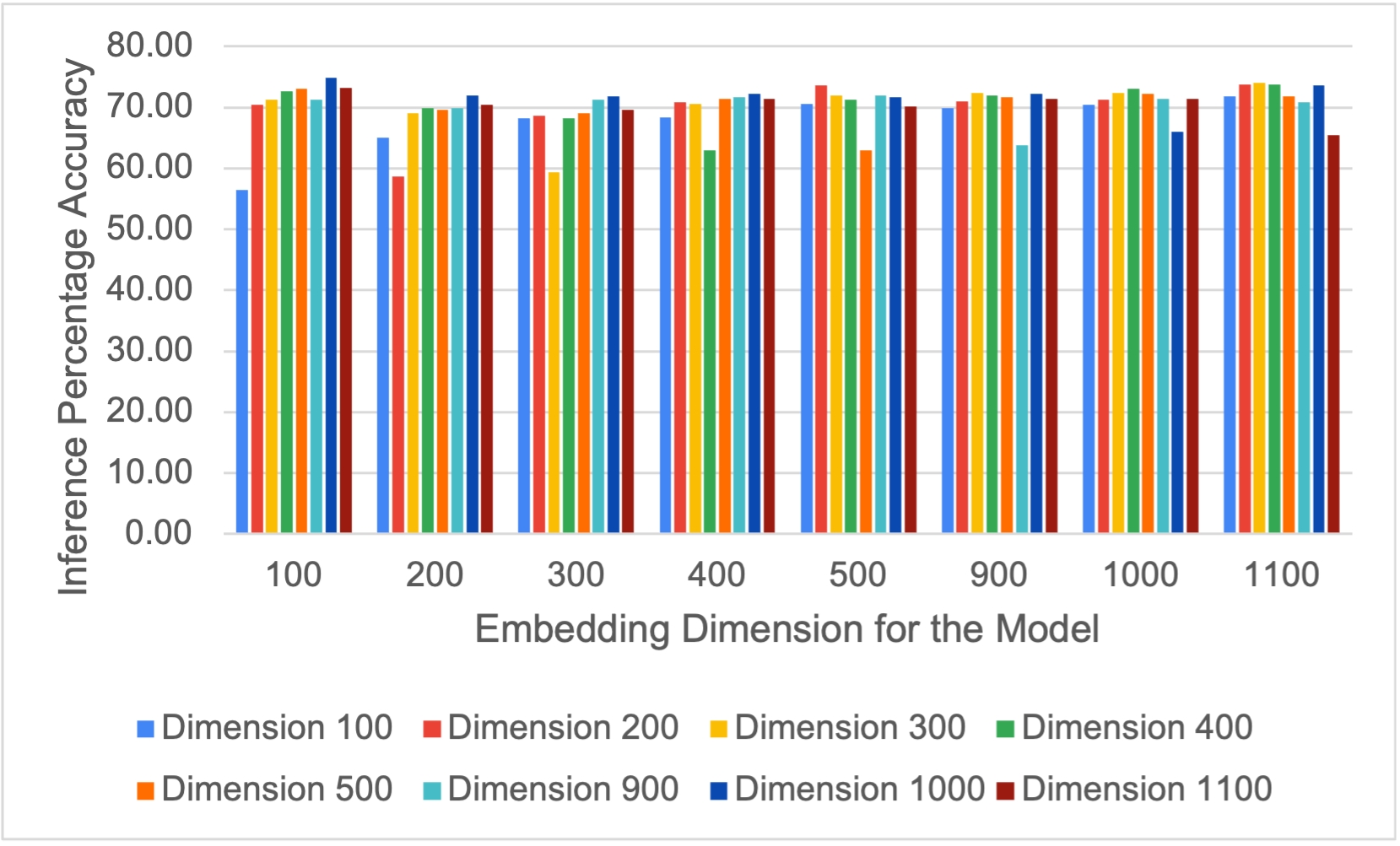}
    \caption{Adversarial vulnerability is sensitive to the embedding dimension, tested on Twitter dataset. Note that for every model, only one attack works (lower bar), when its embedding dimension matches that of the model's inputs. }
    \label{performance_2}
    \end{figure}
    \item Figures \ref{sensitivity_1} and \ref{sensitivity_2} show that the ensemble models work as a defensive counter-measure to adversarial attacks. To briefly explain the plots, each group of bars correspond to a specific type of ensemble model, as mentioned in the labels. Within the groups, the first eight bars from the left are those corresponding to the embedding dimension of the attack, as mentioned in the labels, and the final two on the right are two special cases. The first special case is that of using clean samples instead of an attack, setting a baseline, and the second special case is the best attack scenario, which means the most successful attack within the ones included in the ensemble of models. It is observed in both figures that the performances of the ensemble models are significant improvements over the best adversarial attacks, as they close in on the desired accuracy that is obtained on the clean samples.  
    \begin{figure}[!htbp]
    \centering
    \includegraphics[width=\textwidth]{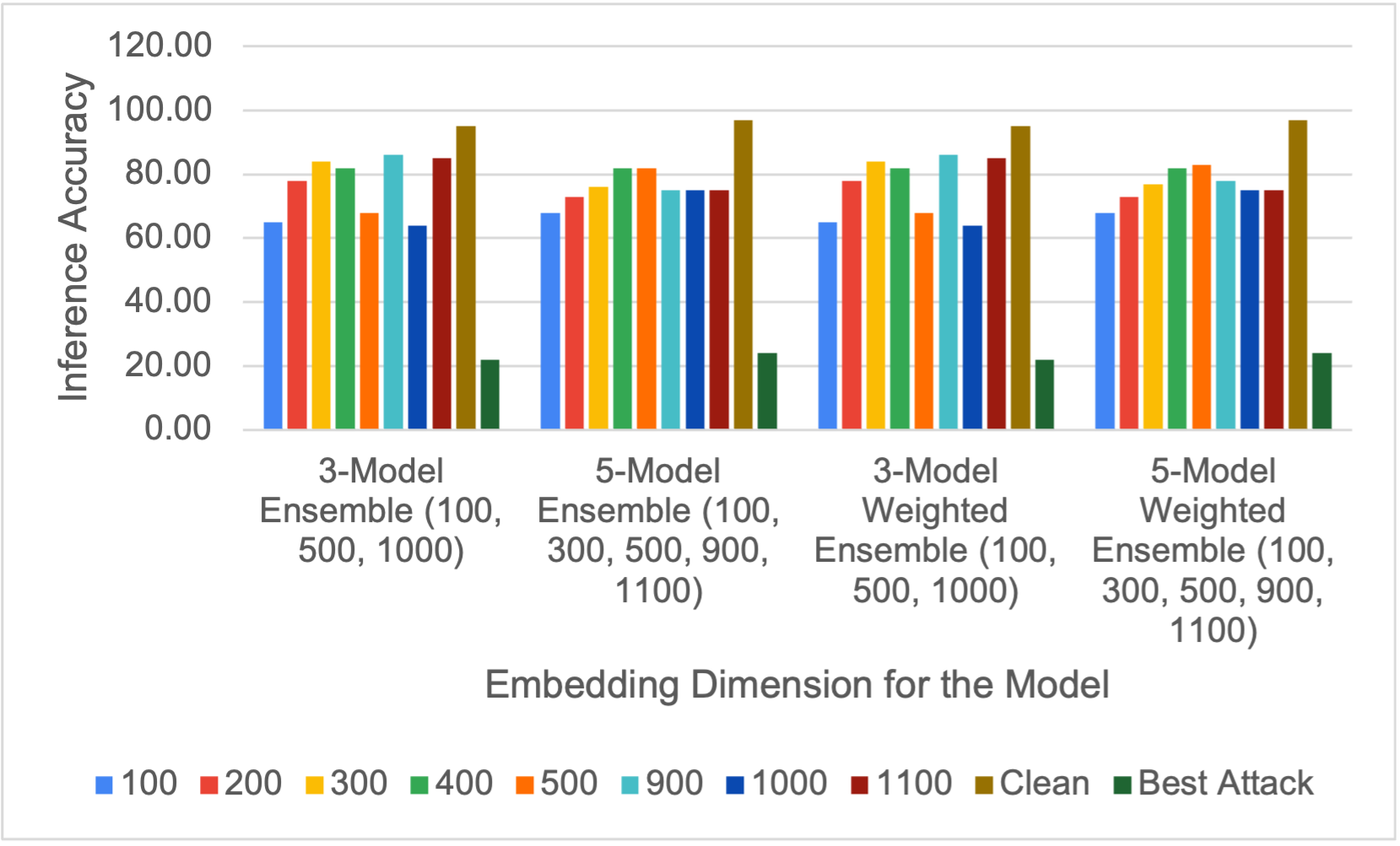}
    \caption{Exploiting the adversarial sensitivity for robustness using ensembles as counter-measure, tested on the IMDB dataset. Note that the performance of the ensemble models is much better (higher bars) across attack modes than an individual model (presented as the best attack in the last green bar).}
    \label{sensitivity_1}
    \end{figure}
    \begin{figure}[!htbp]
    \centering
    \includegraphics[width=\textwidth]{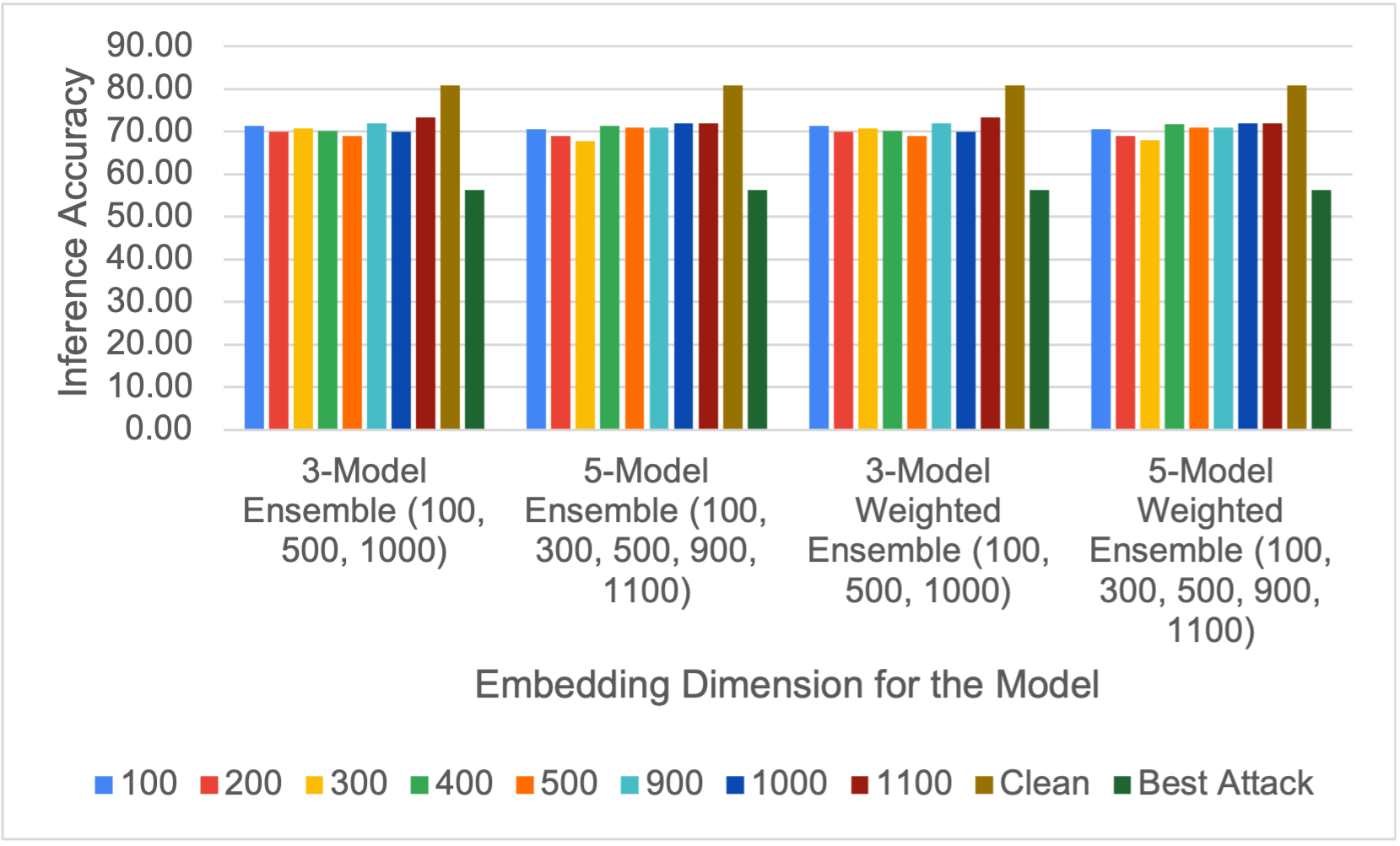}
    \caption{Exploiting the adversarial sensitivity for robustness using ensembles as counter-measure, tested on the Twitter dataset. Note that the performance of the ensemble models is much better (higher bars) across attack modes than an individual model (presented as the best attack in the last green bar).}
    \label{sensitivity_2}
    \end{figure}
    \item The study of the sensitivity of the adversarial samples to its inherent dimensionality reveals that within a very fine margin as well (in the range of 50 as shown in Tables \ref{table1} and \ref{table2}, measured in terms of embedding dimensions and adversarial attack dimensions in the range of 900 through 1100), the success of the attacks are limited to a matching model with the corresponding input embedding dimension. This observation of high sensitivity is well exploited in the defence mechanism involving ensembles. Figures \ref{sensitivity_1} and \ref{sensitivity_2} show that while all the individual models have the weakness for their corresponding specific adversarial samples (in the first 8 bars), the ensemble models are able to overcome this vulnerability. 
    \item The study on the measurement of adversarial perturbation, in Tables \ref{table3} and \ref{table4}, shows that there is a consistent growth in the variability of the measurements of adversarial perturbation as we increase the embedding dimension of the model against which the attack is created, as evident from the distributions and its corresponding coefficients of variation. This study supports the assertion that the fluctuations in adversarial perturbations are correlated with the embedding dimensions. 
\end{itemize}

The experimental results corroborate the claims made earlier regarding the use of ensemble models as a potential adversarial defence against attacks on text classifier models.

\section{Concluding Remarks} \label{conclusions}
Adversarial attacks have long been regarded as one of the primary threats to machine learning algorithms that prevent their widespread adoption. Originally discovered in image classification tasks, adversarial attacks have been heavily studied in the domain of computer vision. In this paper, we investigate adversarial attacks on text classifier problems, try to understand why they occur and use the insights to develop potential counter-measures. 
In particular, we have studied adversarial attacks with respect to the inherent dimensionality of the classification problem, which is often attributed to being one of the primary reasons these adversarial examples exist. We tested different adversarial samples created against models which have varying dimensionality in terms of the input embedding dimension used for training for how sensitive they are to those models as opposed to other models of non-matching dimensionality. 

Our results have shown that there is an extremely high correlation between the success of adversarial attacks on text classifiers with the inherent dimensionality of the models, which is quite unlike what is typically observed in images and other data. This observation also allows us to design potential counter-measures. We have used ensemble models to thwart adversaries, and the results suggest that they have adversarial robustness. 

In the future, we wish to extend this work to other forms of natural language understanding tasks which involve more complex neural architectures, thus increasing the scope of the idea presented in this paper.

\backmatter

\section*{Declarations}

\subsection{Availability of data and materials}
There are two datasets used in this paper. The IMDB dataset used in experiments is the in-built dataset taken from \textit{torchtext.datasets} package in Pytorch. The dataset for Twitter samples is taken from Kaggle. The links to both the datasets are given below:
\begin{itemize}
    \item IMDB dataset: \href{https://pytorch.org/text/stable/datasets.html#imdb}{Link}
    \item Twitter dataset: \href{https://www.kaggle.com/kazanova/sentiment140}{Link}
\end{itemize}

\subsection{Competing interests}
We do not have any conflict of interests to declare.

\subsection{Funding}
This research is funded by the Hardware and Embedded Systems Lab (HESL), School of Computer Science and Engineering, Nanyang Technological University, Singapore. 

\subsection{Authors' contributions}
All authors contributed equally.

\subsection{Acknowledgements}
Not applicable in this case.

\bibliography{sn-bibliography}

\end{document}